\documentclass[conference]{IEEEtran}
\IEEEoverridecommandlockouts
\usepackage{cite}
\usepackage{amsmath,amssymb,amsfonts}
\usepackage{algorithmic}
\usepackage{graphicx}
\usepackage{textcomp}
\usepackage{xcolor}
\usepackage{multirow}
\usepackage{expl3}
\def\BibTeX{{\rm B\kern-.05em{\sc i\kern-.025em b}\kern-.08em
    T\kern-.1667em\lower.7ex\hbox{E}\kern-.125emX}}
\begin{document}

\title{SenTest: Evaluating Robustness of Sentence Encoders\\
\thanks{L3Cube Labs, Pune}
}
\author{\IEEEauthorblockN{Tanmay Chavan}
\IEEEauthorblockA{\textit{Pune Institute of Computer Technology} \\
\textit{L3Cube Labs} \\
Pune, India \\
chavantanmay1402@gmail.com}
\and
\IEEEauthorblockN{Shantanu Patankar}
\IEEEauthorblockA{\textit{Pune Institute of Computer Technology} \\
\textit{L3Cube Labs} \\
Pune, India \\
shantanupatankar2001@gmail.com}
\and
\IEEEauthorblockN{Aditya Kane}
\IEEEauthorblockA{\textit{Pune Institute of Computer Technology} \\
\textit{L3Cube Labs} \\
Pune, India \\
adityakane1@gmail.com}
\and
\IEEEauthorblockN{Omkar Gokhale}
\IEEEauthorblockA{\textit{Pune Institute of Computer Technology} \\
\textit{L3Cube Labs} \\
Pune, India \\
omkargokhale2001@gmail.com
}
\and
\IEEEauthorblockN{Geetanjali Kale}
\IEEEauthorblockA{\textit{Pune Institute of Computer Technology} \\
Pune, India \\
gvkale@pict.edu
 }
\and
\IEEEauthorblockN{Raviraj Joshi}
\IEEEauthorblockA{\textit{Indian Institute of Technology Madras} \\
\textit{L3Cube Labs} \\
Pune, India \\
ravirajoshi@gmail.com}
}

\maketitle

\begin{abstract}

Over the past few years, Natural Language Processing has gained importance due to its empirical advantages in real-world scenarios. Moreover, the advent of transformer-based architectures has made it easy to deploy language models at scale. This comes with an inherent problem of handling noise in the input data at inference time. In other words, it has become increasingly important for models to become more and more robust. Contrastive learning has proven to be an effective method for pre-training models using weakly labeled data in the vision domain. Sentence transformers are the NLP counterparts to this architecture, and have been growing in popularity due to their rich and effective sentence representations. Having effective sentence representations is paramount in multiple tasks, such as information retrieval, retrieval augmented generation (RAG), and sentence comparison. Keeping in mind the deployability factor of transformers, evaluating the robustness of sentence transformers is of utmost importance. This work focuses on evaluating the robustness of the sentence encoders. We employ several adversarial attacks to evaluate its robustness. This system uses character-level attacks in the form of random character substitution, word-level attacks in the form of synonym replacement, and sentence-level attacks in the form of intra-sentence word order shuffling. The results of the experiments strongly undermine the robustness of sentence encoders. The models produce significantly different predictions as well as embeddings on perturbed datasets. The accuracy of the models can fall up to 15 percent on perturbed datasets as compared to unperturbed datasets. Furthermore, the experiments demonstrate that these embeddings does capture the semantic and syntactic structure (sentence order) of sentences. However, existing supervised classification strategies fail to leverage this information, and merely function as n-gram detectors.
\end{abstract}


\section{Introduction}
Most of the human interaction over the internet occurs via the medium of text. Thus, it has become necessary to extensively study this modality of communication. Natural Language Processing is the study of the language used by humans for day-to-day tasks and more. Natural Language Processing (NLP) has gained importance over the past years, mainly due to its wide applications to other industries and fields of study.

The application of deep learning in NLP started in 2000 in the paper \cite{bengio2000neural} where feed-forward neural networks were used to predict a word using n previous words. This was followed by the use of multi-task learning in \cite{collobert2008unified}. The use of word embeddings in NLP tasks was introduced by \cite{mikolov2013distributed} in 2013. The use of the sequence-to-sequence models in NLP was introduced in \cite{sutskever2014sequence} where an encoder-decoder architecture was used for NLP tasks. Earlier, convolutional layers were used in the majority of NLP tasks like Sentiment Analysis, Machine Translation, and Question answering. However, after the discovery of RNNs and their more optimized versions, such as Long Short-Term Memory (LSTMs) and gated recurrent unit (GRU), the focus on CNNs shifted. Since then, the advances in deep learning, and the availability of computing have made it increasingly easier to study and model these interactions.  After the seminal work, "Attention is All You Need" in 2017, transformers have steadily grown in importance and have lately expanded to other fields of artificial like computer vision and graph networks.


Large Language Models (LLMs) have now gained mainstream importance due to their efficacy in solving Natural Language Processing (NLP) problems. Since the advent of transformers \cite{vaswani2017attention, devlin-etal-2019-bert, zhuang-etal-2021-robustly}, many practitioners have moved to transformers as the default solutions for a wide variety of NLP tasks. Moreover, large-scale pretraining has made few-shot and zero-shot approaches 
 a viable alternative for tasks that traditionally required a considerable amount of supervisory signal. Contrastive self-supervised training algorithms are well-known and widely used for pre-training models on large corpora of unlabelled data. Their prowess has been demonstrated in the vision domain, where works like DiNO, SEER, SimCLR and many more show tremendous potential for self-supervised pre-training.

Sentence encoders are the next major advancement in transformers. Unlike transformers, which are trained using the Masked Language Modelling objective, sentence encoders are trained using the contrastive similarity objective. This makes them quantifiably better than MLM-trained transformers, as they gain more semantic information than their predecessors \cite{cer-etal-2018-universal}.  Large corpora of data, availability of large amounts of computing and storage, and ease of access to these resources have made sentence encoders an increasingly popular choice for many tasks. Sentence encoders use the well-known Siamese network architecture. These models have shown empirically better results on a multitude of downstream tasks.

With these rapid advancements and the introduction of high-capacity models, their deployability remains a burning question. Robustness is defined as the attribute of the model to give consistent results even when provided with imperfect inputs or inputs with perturbations. Robustness testing has gained paramount importance as transformer-based systems become increasingly common. This paper aims to evaluate the robustness of sentence encoders against perturbed inputs. We employ several types of perturbations on input data and record their results and outputs generated by the models. We try out perturbations at character-level, word-level, as well as sentence-level. Then we compare these outputs using different methods and measure the robustness of the models against the said perturbations. These comparisons allow us to gain greater insights into the robustness of sentence encoders and highlight their limitations.



\begin{figure*}[h]
  \centering
  \includegraphics[width=\textwidth]{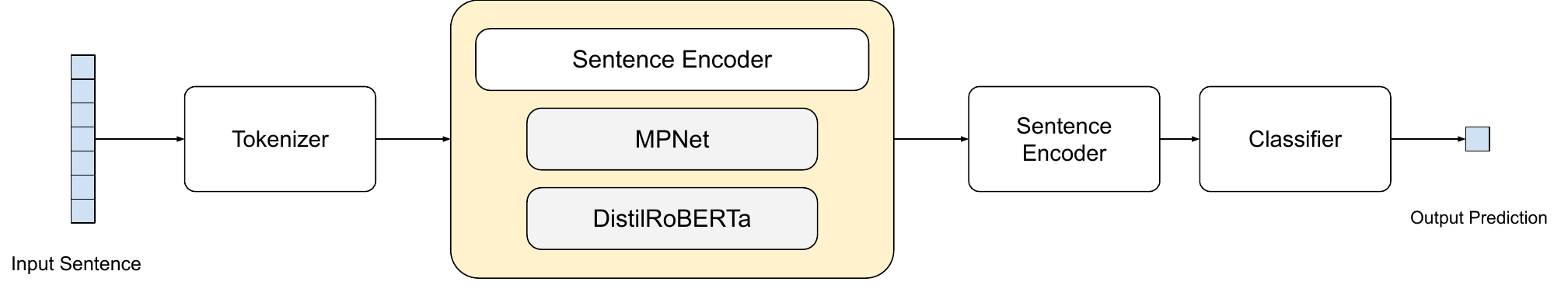}
  \caption{System architecture}
  \label{fig:system_overview}
\end{figure*}

Pre-training of Deep Bidirectional Transformers for Language Understanding (BERT) \cite{devlin-etal-2019-bert} was the major breakthrough that enabled massive developments in the field of Natural Language Processing. It has been the go-to model architecture for a long time. However, BERT can only find word-level encodings which are not always useful in capturing the overall context of the sentence. Sentence encoders were developed to overcome this problem. In the paper, Sentence-BERT: Sentence encodings using Siamese BERT-Networks \cite{reimers-gurevych-2019-sentence}, a new Siamese network-based model called S-BERT is introduced in which parallel, similarity-based training of sentences is used for pre-training is used to capture the context of the sentence. Other papers like Universal Sentence Encoder\cite{cer-etal-2018-universal} have also used a similar approach to obtain sentence-level encodings. This project aims to test the robustness of these models and improve it if required.

Our key observations are as follows.
\begin{itemize}
    \item The pre-trained sentence transformer models are not robust to input perturbations and in some cases, there is a significant drop in test accuracy with these perturbations. Word-level synonym replacement has minimal impact followed by character level changes.
    \item Specifically for sentence-level tasks we use the ShufText method \cite{taware2021shuftext}. With this method, the words in the sentence are randomly shuffled to arrive at a meaningless sentence. On a shuffled test set, there is a significant drop in accuracy. However, this accuracy is still well above the random accuracy indicating decent performance on in-interpretable sentences. Even for humans, it is extremely difficult to identify the target label but these models still understand the underlying label with a decent accuracy. This highlights the fact that even sentence BERT models-based classifiers rely on keywords and just act as n-gram identifiers. 
    \item The sentence representations of the original text and shuffled text are quite close in terms of cosine similarity. This shows that completely changing the sentence structure and meaning has minimal impact on the model-generated representations. The behavior highlights the bag of words nature of sentence representations.
    \item We further experiment to check if the sentence order information is present in the initial sentence embedding. The experiments reveal that the sentence embedding does contain the information however the same is not utilized by the supervised fine-tuned model. Even the cosine similarity metric fails to capture the same.
\end{itemize}

\section{Literature Survey}
LLMs have achieved excellent scores on numerous downstream tasks in Natural Language Processing \cite{liu2019roberta} \cite{zhang2022finemixing} \cite{raffel2020exploring}. These models are trained on enormous amounts of data and are theoretically able to capture the semantics of a language. However, since these models operate as black boxes, little is known about their inner workings. What happens inside the models on a neuron level is unknown. It is observed that such models often fail when small amounts of perturbations are made to the input text. This has given rise to a large amount of research in the evaluation of the robustness of these models. Robustness evaluation includes the organization of adversarial attacks on the input data such that the model's accuracy is reduced significantly. Adversarial attacks are performed by mixing adversarial samples in the input data for a model. These adversarial examples include slight perturbations to the sentence. Each perturbation varies depending on the granularity of the attack. These attacks, or perturbations can be classified into three categories. This section explores the literature on different attacks based on their granularity.

\subsection{Character-level attacks}
The core idea of character-level attacks is to introduce perturbations in characters of the input text to make it marginally different from the original text. The model is then expected to perform almost identically on this perturbed corpus of data. Considerable work has been done in this area, which provides a strong baseline for future research. The simplest approach to this is proposed by \cite{blackboxcharaclevel}. In this work, the authors propose DeepWordBug, a scoring algorithm that determines character-level changes based on a black-box approach. The authors claim that this black-box approach is the fastest approach for character-level adversarial attacks. \cite{gil-etal-2019-white} proposes a white-and-black box approach for generating character-based perturbations. They propose two models -- the first is a white-box model which determines which character change will induce the most change and the second model is a black-box adversarial attacker which generated perturbations for the actual sentence. Another work by \cite{ebrahimi-etal-2018-adversarial} proposes an efficient method for white-box attacks and subsequently proposes an efficient defense strategy that takes only three times the training time. They perform various ablations for the neural machine translation (NMT) task. Their main contribution is a differentiable character edit operations algorithm, which can be optimized using stochastic gradient descent to get the edits with the most impact. Another work pertaining to NMT \cite{noisebreaksnmt} proposes methods to induce natural and synthetic noise in NMT systems to show that they perform poorly in both scenarios. Furthermore, they explore structure invariant word representations to mitigate this problem. \cite{he-etal-2021-model} proposes a two-phase approach for deployable BERT-based systems. The first stage includes querying the black-box victim model to get its predictions and train the attacker model on these predictions. This enables a white-box approach since the attacker model is effectively distilled using the victim's back-box API. The second phase consists of introducing character-level perturbations, namely insertion, deletion, mistypes, incorrect pronunciation-based perturbations, and human behavior-based typographical errors. This work proposes two defense strategies: to assign a temperature to the softmax function used for predictions, and secondly to induce a noise of a given variance in the training data of the model. Lastly, \cite{eger-etal-2019-text} proposes to induce visual modifications that are targeted to hide offensive works, for example, the word "idiot" is spelled as "!di0t" to avoid detection from offensive speech detection systems. The authors show that, unlike humans, machines are extremely susceptible to visual perturbations. They employ various pre-existing defense strategies to reduce this problem.

\subsection{Word-level attacks}
Word-level adversarial attacks largely focus on the insertion and deletion of words in text samples in manners that maximize the impact on the generated encodings and thus the model results. Word-level adversarial attacks are predominantly used as classic tokenizers operate at the word level. Thus word word-level attacks can help us discover the flaws in deep learning models effectively without major modifications in the original text. \cite{https://doi.org/10.48550/arxiv.1907.11932} studies the robustness of BERT models. They provide baselines for robustness as well as deploy word-level adversarial attacks which provides strong result while simultaneously not destroying the semantic content or the grammaticality of the original text. \cite{8682430} presents special types of word-level adversarial attacks that primarily target classifier models. Their gradient-based method of inserting a sequence of tokens in any input has proven to marginally hamper the accuracy of a lot of text classifiers. \cite{yin-etal-2020-robustness} explores the effect human grammatical errors have on large language models. Their studies discovered some grammatical errors have a larger impact than others, and the positions of these errors in the sequence is a deciding factor for their impact. \cite{garg-ramakrishnan-2020-bae} has employed a BERT-based approach to generate adversarial attacks. They mask some words from the input sequence and use a pre-trained large language model to suggest replacements for the masked words. This approach helps insert perturbations without changing the meaning of the sentence. \cite{li-etal-2020-bert-attack} proposes a similar approach where they use a pre-trained BERT model to identify the weak spots within input sequences and insert perturbations at those spots, thus creating better adversarial inputs. \cite{ren-etal-2019-generating} focuses on the problem that most adversarial attacks fail to insert adversarial perturbations in places to cause maximal impact. They propose a novel approach using a greedy algorithm that takes into account the word saliency as well as the classification probability.

\subsection{Sentence-level attacks}
Sentence-level adversarial attacks are perturbations where we modify entire sentences in the provided input text. This may involve addition, deletion, and modification of sentences present in the input. Sentence-level adversarial attacks are less studied than their word-level counterparts as most of the encoders and tokenizers used in natural language processing break down sentences into several tokens. However, the advancements in sentence encoders have ensured a resurged interest in sentence-level adversarial attacks. \cite{lin-etal-2021-using} studies the impact of adversarial attacks on Machine Reading Comprehension models. The authors demonstrate that one adversarial option along with three regular options can significantly hamper the performance of MRC models. \cite{gan-ng-2019-improving} checks the robustness of modern Question Answering (QA) models. It also proposes a new approach of data augmentation to enable the paraphrasing of input data without any direct human involvement. \cite{xu-etal-2021-grey} presents a new grey-box adversarial attack framework meant for sentiment classification. \cite{wang-etal-2020-cat} presents a new framework for adversarial attacks where we can employ perturbations based on controllable attributes. This feature enables us to augment one particular feature of the input data without modifying other features of the text. \cite{https://doi.org/10.48550/arxiv.1710.11342} focuses on the problem of unnatural perturbations that provide unrealistic adversarial attacks that fall short of replicating real-world scenarios. It proposes a new method of generating adversarial perturbations using generative adversarial networks which is able to present more realistic adversarial examples.

\subsection{Multi-level attacks}
The above sections feature individual adversarial attacks. However, a single type of perturbation rarely occurs in a real-world scenario. Most attacks occur as a combination of the types mentioned above. To combat this, multi-level attacks are used to evaluate robustness. \cite{chen2021multi} uses the similarity of sentence embeddings to pick the perturbations that are most similar to the original text. It uses a reinforcement learning-based approach to organize adversarial attacks. Oftentimes, the input data to language models is codemixed, that is the text is a combination of more than one language. Such cases may lead to a drop in the performance of the models. \cite{tan2021code} uses word-level and phrase-level transliterations as perturbations to reduce the accuracy of XLMR on the XNLI data from 79.85\% to 8.18\%.  Perturbations are also organized by using deep learning models. \cite{song2020universal} creates perturbations by using a generator that generates them using the noise obtained from the gradient of the target model's loss function. Adversarial attacks are also organized for domain-specific tasks. The Biomedical BERT-based Adversarial Example Generation \cite{mondal2021bbaeg} creates adversarial samples specifically based on biomedical synonyms. Automatic scoring systems can be easily fooled by students to obtain partial answers, as demonstrated in \cite{ding2020don}. This problem can be solved using adversarial training. Some multi-level attacks are universal in nature. They are agnostic to factors like inputs and models. \cite{wallace2019universal} uses a white box method to generate universal triggers. These triggers work across almost all tasks, models, and datasets. Textbugger\cite{li2018textbugger} is a framework that uses white box and black box attacks to reduce the accuracy of Deep learning-based Text Understanding. Almost all adversarial attacks were black box attacks. However, HotFlip \cite{ebrahimi2017hotflip} is a purely white box method. Hotflip uses character-level flips on the one-hot input representations to significantly reduce the performance of language models. From the literature survey of these multi-level attacks, a significant gap in the number of white-box methods is observed. A combination of both black box and white box methods can be used to yield a maximum decrease in model performance.

\section{Description of Datasets}
This paper presents experiments conducted on four distinct datasets, consisting of two short-text and two long-text datasets. The two long-text datasets used in this study are the BBC News and IMBD datasets, while the short-text datasets are TREC and Emotion. The train and test split numbers are mentioned in Table \ref{tab:data_stat}.

\subsection{TREC Dataset}
The TREC Question Classification dataset consists of 5,500 questions, with 5,000 questions in the training set and 500 questions in the test set. It includes six coarse class labels and 50 fine class labels. The average sentence length of this dataset is 10 words, and its vocabulary size is 8,700.

\begin{table}[]
\begin{tabular}{|l|l|l|l|l|}
\hline
Dataset Name & Average Length & Unique Vocab & Train & Test  \\ \hline
TREC         & 10             & 8700         & 5000  & 500   \\ \hline
Emotion      & 19             & 16196        & 18000 & 2000  \\ \hline
IMDB         & 235            & 9998         & 25000 & 25000 \\ \hline
BBC News     & 421            & 14657        & 1225  & 1000  \\ \hline
\end{tabular}
\caption{Dataset Statistics}
\label{tab:data_stat}
\end{table}

\subsection{Emotion Dataset}
The Emotion Dataset contains a corpus of 20,000 labeled tweets, each labeled with one of the five emotions: sadness, joy, love, anger, fear, and surprise. The dataset has a vocabulary of 16,169 words and an average sentence length of 19 words.

\subsection{IMDB Dataset}
The IMDB dataset comprises 50,000 movie reviews labeled as either positive or negative sentiments. The dataset has been divided into two subsets of 25,000 reviews each for training and testing purposes. It contains a vocabulary of 9,998 unique words, and the average length of each review is 234.75 words. This dataset is commonly used in sentiment analysis research for predicting the sentiment of movie reviews and analyzing trends in movie reviews over time.

\subsection{BBC News Dataset}
The BBC News dataset consists of 2,225 news articles from the BBC News website, each labeled under one of five categories: business, entertainment, politics, sport, or tech. The dataset is split into 1,490 records for training and 735 for testing. The average length of each article is 421 words, and its vocabulary size is 14,657.

\section{Methodology}

\begin{figure*}[h]
  \centering
  \includegraphics[]{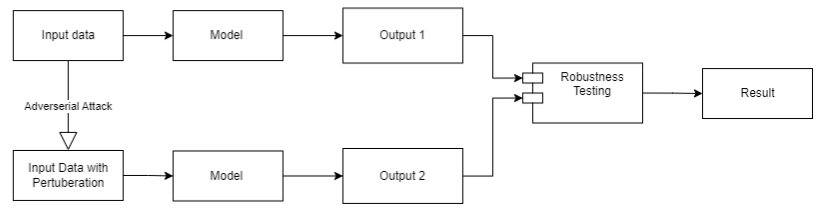}
  \caption{Generalized framework of operation for sentence encoders.}
  \label{fig:System Diagram}
\end{figure*}

This section provides an overview of the system setup that was employed to carry out a series of robustness testing experiments. The main objective of these experiments was to evaluate the ability of two popular models, namely distil-roberta-v1 and mpnet-base-v2, to handle data manipulation and generalize to diverse datasets. To this end, we generated perturbed test sets by implementing three types of attacks: shuffling, character replacement, and synonym replacement. These attacks were intended to mimic real-world scenarios where the input data may be subject to various forms of manipulation.

To conduct the experiments, we first fine-tuned the models on the corresponding training sets of four diverse datasets. Subsequently, we used these models to make predictions for both the original and perturbed test sets, and we recorded the performance drop on the perturbed test set using various evaluation metrics.

In the following sections of the paper, we analyze the results of these experiments in order to gain insights into the robustness of the models. In particular, we use a technique that enables us to evaluate the models' ability to handle data manipulation and generalize to diverse datasets. By comparing the performance of the models on the original and perturbed test sets, we are able to identify potential vulnerabilities and limitations of the models.

    \section*{Text Shuffling Algorithm}

    \noindent \textbf{Input:} \emph{text} (a string of words)

    \noindent \textbf{Output:} \emph{shuffledText} (the input text with words cleaned and shuffled)

    \begin{enumerate}
        \item Clean the input text:
        \begin{itemize}
            \item Remove any leading or trailing whitespace.
            \item Remove punctuation marks.
            \item Convert the text to lowercase.
        \end{itemize}

        \item Split the cleaned text into a sequence of words.

        \item Shuffle the sequence of words.

        \item Concatenate the shuffled words into the shuffled text.

        \item Output the shuffled text.
    \end{enumerate}

    \section*{Character Replacement Algorithm}

    \noindent \textbf{Input:} \emph{sentence} (a string of words)

    \noindent \textbf{Output:} \emph{modifiedSentence} (the input sentence with randomly replaced characters)

    \begin{enumerate}
        \item Clean the input sentence: remove leading/trailing whitespace, punctuation marks, and convert to lowercase.

        \item Split the cleaned sentence into words.

        \item Choose 5 percent of the words randomly.

        \item For each selected word, replace random characters with adjacent keyboard letters.

        \item Concatenate the modified words into the modified sentence.

        \item Output the modified sentence.
    \end{enumerate}

    \section*{Word Replacement Algorithm}

    \noindent \textbf{Input:} \emph{sentence} (a string of words)

    \noindent \textbf{Output:} \emph{modifiedSentence} (the input sentence with words replaced by their synonyms)

    \begin{enumerate}
        \item Split the input sentence into words.

        \item Choose 20 percent of the words randomly.

        \item For each selected word, replace it with a synonym.

        \item Concatenate the modified words into the modified sentence.

        \item Output the modified sentence.
    \end{enumerate}

    \noindent

\section{Experiments}


In order to quantify the robustness of sentence encoders, our experiments constitute performing various perturbations to the input sentence. In our experiments, we have three types of perturbations, namely sentence level, word level, and character level.


\subsection{Sentence level perturbation}
Sentence-level perturbations aim to observe and quantify the effect of introducing perturbations on the sentence level, that is to the entire sentence. In sentence-level perturbation, the perturbation is applied to the given input sentence as a whole. More specifically, in our case, the given sentence is shuffled. In the experiment of sentence level perturbation, we calculate accuracy on the clean dataset. After this, we shuffle the sentences in the testing dataset and calculate accuracy and other metrics on this shuffled counterpart. 

\begin{figure}[h]
  \centering
  \includegraphics[width=0.5\textwidth]{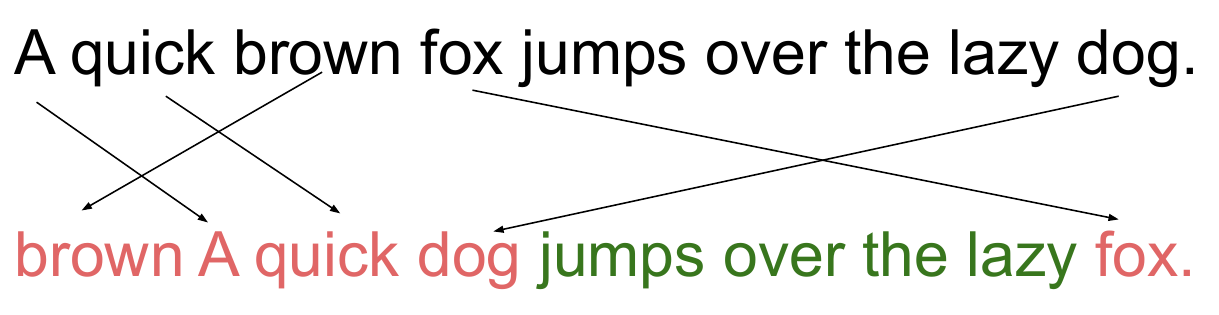}
  \caption{Sentence level perturbation}
  \label{fig:shuffle_perturbation}
\end{figure}

Here, the shuffling of the sentences is the worst-case analysis of an adversarial attack. We observe the effect of complete shuffling of the dataset on accuracy and other metrics and thus deduce that the observed effect is indeed the upper bound of the difference of metrics among the clean and shuffled data. Any lower degree of shuffling will invariably lead to a lesser delta in metrics between the clean and shuffled data. This method is illustrated in Fig. \ref{fig:shuffle_perturbation}.

\subsection{Word level perturbation}

Word level perturbation is aimed to measure the effects of passing perturbed examples to the model and to see the effect of perturbations on the word level. This means that the individual words in the sentence are changed to demonstrate such perturbation. In our experiments, we replace adjectives and adverbs in the given sentence with their synonyms. The specific choice of replacing adjectives and adverbs stemmed from the fact that changing proper nouns and verbs may completely change the meaning and are unlikely to happen in a real-world scenario. 

\begin{table*}
\centering

\begin{tabular}{|c|c|cccc|cccc|}
\hline
\multirow{2}{*}{\textbf{Metric}}     & \multirow{2}{*}{\textbf{Type}} & \multicolumn{4}{c|}{\textbf{MPNet}}                                                                                         & \multicolumn{4}{c|}{\textbf{DistilRoberta}}                                                                                      \\ \cline{3-10} 
                                     &                                & \multicolumn{1}{c|}{\textbf{TREC}} & \multicolumn{1}{c|}{\textbf{Emotion}} & \multicolumn{1}{c|}{\textbf{BBC News}} & \textbf{IMDB} & \multicolumn{1}{c|}{\textbf{TREC}} & \multicolumn{1}{c|}{\textbf{Emotion}} & \multicolumn{1}{c|}{\textbf{BBC News}} & \textbf{IMDB}   \\ \hline
\multirow{4}{*}{\textbf{Accuracy}}   & Clean predictions              & \multicolumn{1}{c|}{0.966}         & \multicolumn{1}{c|}{0.933}            & \multicolumn{1}{c|}{0.973}             & 0.938         & \multicolumn{1}{c|}{0.972}         & \multicolumn{1}{c|}{0.927}            & \multicolumn{1}{c|}{0.976}             & 0.920         \\ \cline{2-10} 
                                     & Shuffled predictions           & \multicolumn{1}{c|}{0.768}         & \multicolumn{1}{c|}{0.818}            & \multicolumn{1}{c|}{0.97}              & 0.855         & \multicolumn{1}{c|}{0.762}         & \multicolumn{1}{c|}{0.809}            & \multicolumn{1}{c|}{0.963}             & 0.838         \\ \cline{2-10} 
                                     & Keyboard predictions           & \multicolumn{1}{c|}{0.8}           & \multicolumn{1}{c|}{0.755}            & \multicolumn{1}{c|}{0.97}              & 0.911         & \multicolumn{1}{c|}{0.85}          & \multicolumn{1}{c|}{0.744}            & \multicolumn{1}{c|}{0.966}             & 0.873         \\ \cline{2-10} 
                                     & Synonym predictions            & \multicolumn{1}{c|}{0.954}         & \multicolumn{1}{c|}{0.787}            & \multicolumn{1}{c|}{0.976}             & 0.932         & \multicolumn{1}{c|}{0.96}          & \multicolumn{1}{c|}{0.767}            & \multicolumn{1}{c|}{0.96}              & 0.913         \\ \hline
\multirow{3}{*}{\textbf{Percentage}} & Clean v/s Shuffled             & \multicolumn{1}{c|}{0.778}         & \multicolumn{1}{c|}{0.854}            & \multicolumn{1}{c|}{0.984}             & 0.867         & \multicolumn{1}{c|}{0.764}         & \multicolumn{1}{c|}{0.832}            & \multicolumn{1}{c|}{0.971}             & 0.859         \\ \cline{2-10} 
                                     & Clean v/s Keyboard             & \multicolumn{1}{c|}{0.812}         & \multicolumn{1}{c|}{0.788}            & \multicolumn{1}{c|}{0.985}             & 0.936         & \multicolumn{1}{c|}{0.858}         & \multicolumn{1}{c|}{0.777}            & \multicolumn{1}{c|}{0.974}             & 0.907         \\ \cline{2-10} 
                                     & Clean v/s Synonym              & \multicolumn{1}{c|}{0.982}         & \multicolumn{1}{c|}{0.826}            & \multicolumn{1}{c|}{0.989}             & 0.980         & \multicolumn{1}{c|}{0.98}          & \multicolumn{1}{c|}{0.800}            & \multicolumn{1}{c|}{0.966}             & 0.970         \\ \hline
\multirow{3}{*}{\textbf{Cosine}}     & Shuffled v/s Clean             & \multicolumn{1}{c|}{0.749}         & \multicolumn{1}{c|}{0.848}            & \multicolumn{1}{c|}{0.98}              & 0.723         & \multicolumn{1}{c|}{0.740}         & \multicolumn{1}{c|}{0.828}            & \multicolumn{1}{c|}{0.966}             & 0.744         \\ \cline{2-10} 
                                     & Keyboard v/s Clean             & \multicolumn{1}{c|}{0.795}         & \multicolumn{1}{c|}{0.757}            & \multicolumn{1}{c|}{0.982}             & 0.880         & \multicolumn{1}{c|}{0.840}         & \multicolumn{1}{c|}{0.753}            & \multicolumn{1}{c|}{0.977}             & 0.850         \\ \cline{2-10} 
                                     & Synonym v/s Clean              & \multicolumn{1}{c|}{0.981}         & \multicolumn{1}{c|}{0.812}            & \multicolumn{1}{c|}{0.99}              & 0.973         & \multicolumn{1}{c|}{0.977}         & \multicolumn{1}{c|}{0.786}            & \multicolumn{1}{c|}{0.969}             & 0.969         \\ \hline
\end{tabular}
  \caption{F1 scores of models when tested on perturbed data. This table compares the accuracy of MPNet and DistilRoberta models on perturbed and unperturbed datasets. The percentage rows indicate \% overlap of labels predicted on clean and perturbed test set. Similarly cosine rows indicate the avg cosine similarity score between those test sets. Note that cosine similarity score and percentage overlap is significant even for shuffled vs clean setting.}
  \label{tab:final_results}
\end{table*}

\begin{figure}[h]
  \centering
  \includegraphics[width=0.5\textwidth]{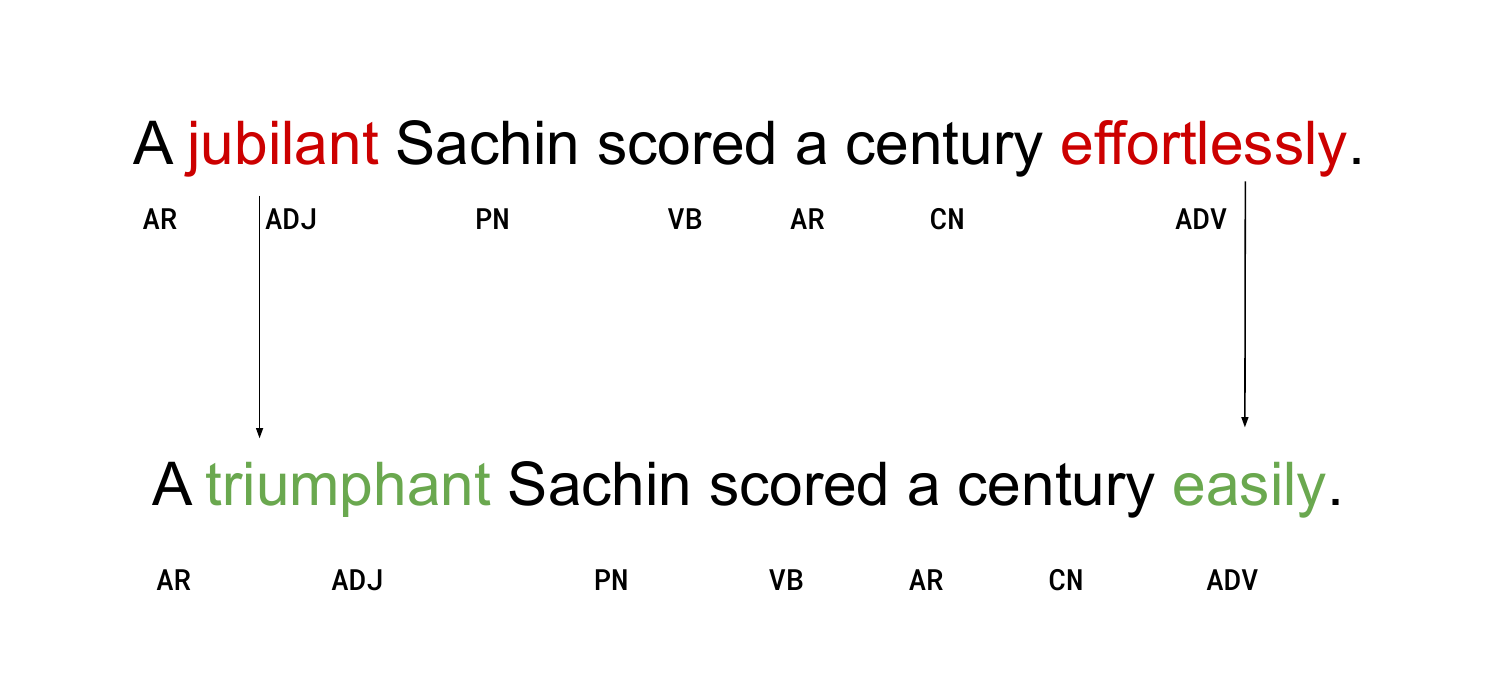}
  \caption{Word level perturbation}
  \label{fig:synonym_perturbation}
\end{figure}

Our method uses the thesaurus to fetch synonyms of the given adjective or adverb. We replace 20\% of the words in a given sentence with their synonyms using this method. Due to the design of this method, it does result in a relatively small difference between the performance of the trained model's clean and perturbed dataset. However, we claim that this is a probable scenario in real-world systems, since non-native speakers may use the irrelevant synonyms of a given word which may change the meaning of the sentence, thus potentially changing model outputs. This method is illustrated in Fig. \ref{fig:synonym_perturbation}.

\subsection{Character level perturbation}

Character level perturbation quantifies the effect of testing on samples that are perturbed on character level. Our system implements this through the replacement of a number of characters in a given sentence. Some characters in the given sentence are replaced by characters in the vicinity of the replaced character on the usual QWERTY keyboard. This experiment is the closest to the real-world scenario since most spelling errata occur by the incorrect replacement of a character with a character in its proximity on the keyboard.

\begin{figure}[h]
  \centering
  \includegraphics[width=0.5\textwidth]{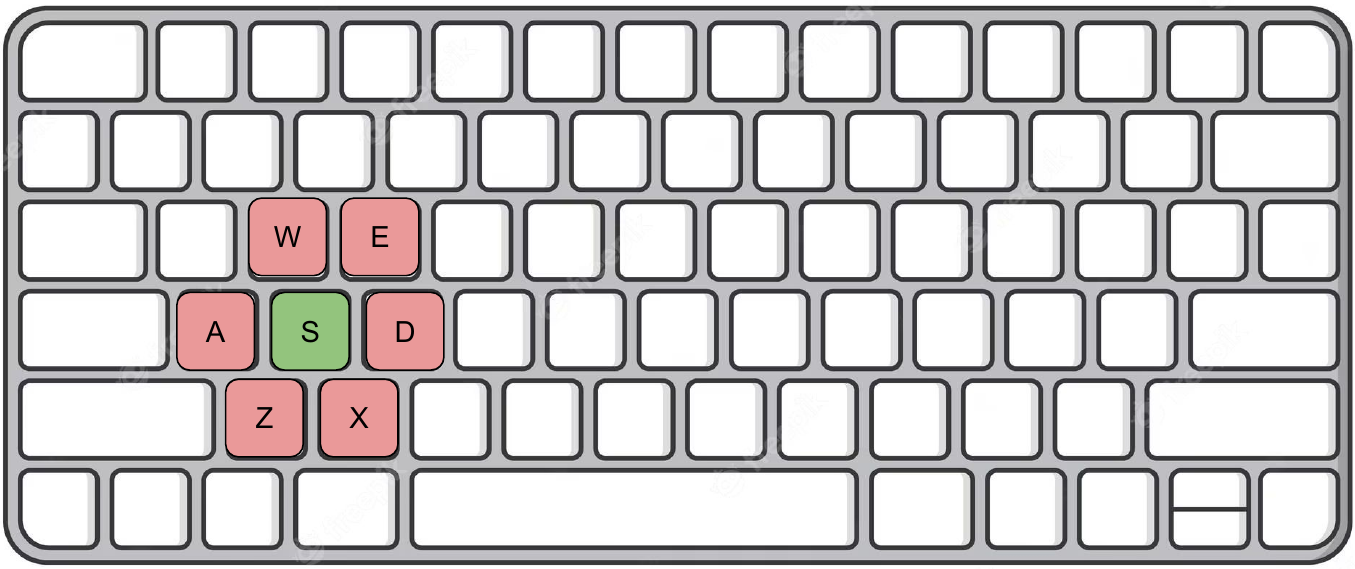}
  \caption{Character level perturbation}
  \label{fig:keyboard_perturbation}
\end{figure}

This method uses a dictionary to determine the replacement character for a given character. Each letter in the dictionary points to multiple letters which are in the vicinity of the given letter on the keyboard. The algorithm then chooses one letter of these letters randomly and replaces it in the given sentence. To make this as close as possible to the real world, we replace 5\% of the characters since the errors that can be attributed to this in a real-world scenario are significant but not higher than other errata commonly occurring in the real-world text. This method is shown in Fig. \ref{fig:keyboard_perturbation}.

\section{Results and observations}

This section summarizes and presents the results of the experiments. Our system generates outputs on perturbed as well as unperturbed datasets. A comparison of the outputs is conducted and the difference is evaluated.

\subsection{Shuffling Experiments}
The shuffling experiments show us that there is a significant reduction in accuracy between the unshuffled data and the shuffled data. However, the shuffled test set accuracy is well above random indicating that these classifiers act as n-gram classifiers and do not pay attention to the syntactic or semantic structure. One interesting phenomenon observed here is that the cosine similarity between the shuffled and unshuffled embeddings is significant enough, especially considering that the words have been shuffled. This might be the reason why accuracy on the shuffled test sets is well above random. The final results are elaborated in Table \ref{tab:final_results}. To further study this phenomenon we devise a simple experiment:\\
\textbf{1. Shuffling classifier experiment:}\\
We mix shuffled and unshuffled sentences in a 1:1 ratio. We label the shuffled sentences as 1 and the unshuffled sentences as 0. We then obtain the embeddings of these sentences using Sbert models and train a binary classifier to determine if a sentence is shuffled or not using its embedding. The results of the shuffled Classifier experiment are showcased in Table \ref{tab:shuff-exp}. We obtain an accuracy of 90\% using a standard ANN-based classifier and an accuracy of 80\% even with simple models like KNN. This experiment shows that the embeddings obtained from Sbert have significant information regarding the structure of the sentence.
\begin{table}[h]
\centering
\begin{tabular}{|c|c|c|c|}
\hline
\textbf{Dataset}                                                                         & \textbf{Embeddings}             & \textbf{Model} & \textbf{Accuracy} \\ \hline
\multirow{4}{*}{\textbf{\begin{tabular}[c]{@{}c@{}}TREC \\ \\ Dataset\end{tabular}}}    & \multirow{2}{*}{mpnet}          & Single layer Neural Network            & 89.6\%              \\ \cline{3-4} 
                                                                                         &                                 & KNN            & 77\%              \\ \cline{2-4} 
                                                                                         & \multirow{2}{*}{distil-roberta} & Single layer Neural Network            & 89.9\%              \\ \cline{3-4} 
                                                                                         &                                 & KNN            & 74.1\%              \\ \hline
\multirow{4}{*}{\textbf{\begin{tabular}[c]{@{}c@{}}BBC news \\ \\ dataset\end{tabular}}} & \multirow{2}{*}{mpnet}          & Single layer Neural Network            & 99.8\%              \\ \cline{3-4} 
                                                                                         &                                 & KNN            & 89.85\%              \\ \cline{2-4} 
                                                                                         & \multirow{2}{*}{distil-roberta} & Single layer Neural Network            & 99.7\%              \\ \cline{3-4} 
                                                                                         &                                 & KNN            & 90.25\%              \\ \hline
\end{tabular}
\caption{Results of the shuffling classifier experiments.}
\label{tab:shuff-exp}
\end{table}

As shown in the above experiment, the sentence embeddings do contain information about the structure of the sentence. 
However, the embeddings still have high cosine similarity and there is considerable overlap in labels predicted on actual and shuffled test sets (Table \ref{tab:final_results}).
We hypothesize that although there is some sentence structure-related information in the embeddings, it is either insufficient or the task-specific model is unable to leverage it to produce good results.

Furthermore, it can be deduced from the above experiment that the structural information is present in the embeddings produced by the sentence encoders, however, the linear classifier is unable to grasp the structural information. This indeed leads to the shuffled and clean sentences producing a relatively small difference in classification metrics.

\subsection{Character Replacement}
The character replacement experiment yields similar results to the shuffled experiments. There is a significant reduction in the accuracy of the text classification models. This is because the replaced characters often generate words that are not in the vocabulary of the tokenizer. This leads to the tokenizer labeling these words as unknown tokens. Due to the increase in the number of unknown tokens, we observe a reduction in model performance. 
\subsection{Synonym Replacement}
In the case of the synonym experiment, we observe that there is no significant reduction in performance. We suspect that the marginal difference observed is due to some error in the synonym replacement model. The gensim model often replaces words with close words based on embeddings rather than context. This may have caused a minor drop in performance. Out of all the datasets, the emotion dataset has a significant drop in performance. Further investigation is needed to explain this behavior.

\section{Conclusion}

This work explores the current state-of-the-art large language models prevalent in the domain. We categorize the robustness attacks based on character, word, sentence, or multi-level attacks and explore each of them in detail. We conducted several experiments such as character replacement, synonym replacement, and word order shuffling on the evaluation dataset. We evaluate differences in model predictions and outputs between clean datasets and perturbed datasets. We observe that the performance and outputs of the model indicate that the models perform poorly on perturbed datasets. Our experiments prove that sentence-encoder-based models lack the necessary robustness. Proposing ideas for mitigating the lack of robustness will prove to be crucial for the adoption of sentence encoders in real-world deployment.

\section*{Acknowledgements}
This work was done under the L3Cube Pune mentorship
program. The problem statement and ideas presented in this work originated from L3Cube and its mentors. We would like to express our gratitude towards
our mentors at L3Cube for their continuous support and
encouragement.

\bibliography{ref, anthology}
\bibliographystyle{IEEEtran}

\end{document}